\documentclass[10pt, twocolumn, letterpaper]{article}

\usepackage{cvpr}
\usepackage{times}
\usepackage{epsfig}
\usepackage{graphicx}
\usepackage{amsmath}
\usepackage{amssymb}
\usepackage{subfig}
\usepackage{algorithm}
\makeatletter
\@namedef{ver@everyshi.sty}{}
\makeatother
\usepackage{tikz, pgfplots}\pgfplotsset{compat=1.9}
\usepackage[noend]{algpseudocode}

\DeclareMathOperator*{\argmin}{arg\,min}

\usepackage[pagebackref=true,breaklinks=true,letterpaper=true,colorlinks,bookmarks=false]{hyperref}

\cvprfinalcopy 

\def\cvprPaperID{8} 

\ifcvprfinal\fi
\begin{document}

\newcommand{\CS}[1]{{\color{orange}\textbf{[CS:} #1\textbf{]}}}
\newcommand{\TODO}[1]{{\color{red}#1}}
\newcommand{\AZ}[1]{{\color{cyan}\textbf{[AZ:} #1\textbf{]}}}

\title{Content Adaptive Optimization for Neural Image Compression}

\author{Joaquim Campos \quad Simon Meierhans \quad Abdelaziz Djelouah \quad Christopher Schroers \\ 
DisneyResearch$\mid$Studios\\
{\tt\small abdelaziz.djelouah@disney.com \quad christopher.schroers@disney.com}
}

\maketitle
 
\begin{abstract}
The field of neural image compression has witnessed exciting progress as recently proposed architectures already surpass the established transform coding based approaches. While, so far, research has mainly focused on architecture and model improvements,
in this work we explore content adaptive optimization. To this end, we introduce an iterative procedure which adapts the latent representation to the specific content we wish to compress while keeping the parameters of the network and the predictive model fixed. Our experiments show that this allows for an overall increase in rate-distortion performance, independently of the specific architecture used. Furthermore, we also evaluate this strategy in the context of adapting a pre-trained network to other content that is different in visual appearance or resolution. Here, our experiments show that our adaptation strategy can largely close the gap as compared to models specifically trained for the given content while having the benefit that no additional data in the form of model parameter updates has to be transmitted.
\end{abstract}

\section{Introduction}
\pgfplotsset{compat=1.3}
\pgfplotsset{grid style={color=black!15, },} 
\pgfplotsset{every axis x grid/.style={draw opacity=0}}

\pgfplotsset{
	axis x line=bottom,
	axis y line=left,
	axis line style={draw=none},
	tick style={draw=none},
	grid,
	xmin=0, xmax=1,
	xlabel=Rate (bpp),
	legend style={at={(0.96,0.04)},anchor=south east},
	legend cell align={left},
	label style={font=\tiny},
	legend style={font=\tiny},
	legend style={draw=gray},
	tick label style={font=\tiny},
	width=0.725\columnwidth,
	legend style = {font = {\fontsize{12 pt}{12 pt}\selectfont}},  
}
\pgfplotsset{
	every axis plot/.append style={line width=1pt, cap=round, smooth},
}
\pgfplotsset{
	compat=newest,
	/pgfplots/legend image code/.code={%
		\draw[mark repeat=2,mark phase=2,#1] 
		plot coordinates {
			(0cm,0cm) 
			(0.25cm,0cm)
			(0.5cm,0cm)
		};
	},
}

\definecolor{myBlue}{RGB}{41,77,165}
\definecolor{myBlueLight}{RGB}{164,185,226}
\definecolor{myOrange}{RGB}{231,116,46 }
\definecolor{myOrangeLight}{RGB}{239,177,105}
\definecolor{myGreen}{RGB}{78,145,41}
\definecolor{myGreenLight}{RGB}{160,219,207}
\definecolor{myRed}{RGB}{181,45,38}
\definecolor{myRedLight}{RGB}{233,138,133}
\definecolor{myMagenta}{RGB}{233,0,233}

\newcommand{\bpg}{myBlue}
\newcommand{\jpg}{myRedLight}
\newcommand{\jpegnew}{myRedLight}
\newcommand{\webp}{myGreenLight}
\newcommand{\full}{myRed}
\newcommand{\simple}{myGreen}

\begin{figure}[t]
	\hspace{-0.2cm}
	\begin{tikzpicture}
	\begin{axis}[ylabel=Distortion (PSNR),ymin=32, ymax=41,xmin=0.3, xmax=1.15,ytick={32, 34,...,44}, scale=1.45,legend style={at={(0.97,.45)},anchor=north east,nodes={scale=0.6, transform shape}, row sep=0.0pt} ]
	\addplot[draw=\full, dashed, line width=0.5pt, mark=*, mark options={scale=0.5,\full,solid}] table [x=bpp, y=psnr, col sep=comma] {evaluation/Tecnick/full.csv};
	\addlegendentry{full~\cite{balle2018full}}
	\addplot[draw=\full, line width=0.5pt,  mark=*, mark options={scale=0.5,\full,solid}] table [x=bpp, y=psnr, col sep=comma] {evaluation/Tecnick/full+adapt.csv};
	\addlegendentry{full+adapt}
	\addplot[draw=\simple, dashed, line width=0.5pt,  mark=*, mark options={scale=0.5,\simple,solid}] table [x=bpp, y=psnr, col sep=comma] {evaluation/Tecnick/simple.csv};
	\addlegendentry{simple~\cite{balle2017endtoend}}
	\addplot[draw=\simple, line width=0.5pt, mark=*, mark options={scale=0.5,\simple,solid}] table [x=bpp, y=psnr, col sep=comma] {evaluation/Tecnick/simple+adapt.csv};
	\addlegendentry{simple+adapt}

	\addplot[draw=\bpg, line width=0.5pt] table[x=bpp, y=psnr, col sep=comma] {evaluation/Tecnick/bpg.csv};
	\addlegendentry{BPG}
	\addplot[draw=\jpegnew, line width=0.5pt] table [x=bpp, y=psnr, col sep=comma] {evaluation/Tecnick/jpeg2000.csv};
	\addlegendentry{JPG2K}
	\addplot[draw=\webp, line width=0.5pt] table [x=bpp, y=psnr, col sep=comma] {evaluation/Tecnick/webp.csv};
	\addlegendentry{Webp}
	\end{axis}
	\end{tikzpicture}
	\caption{\textbf{Latent Adaptation.} 
		The evaluation on the Tecnick dataset~\cite{tecnick} shows that
		per image latent representation adaptation is complementary 
		to existing neural image compression methods. It allows to improve rate-distortion performance while keeping the neural network and the computing time on the decoder side unchanged.}
	\label{fig:teaser}
	\vspace{-0.4cm}
\end{figure}
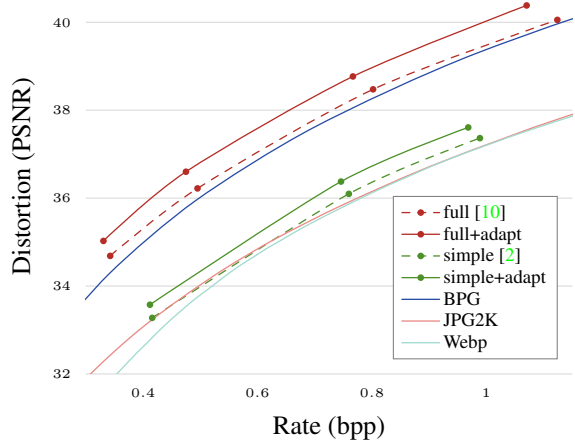

The share of video content in today's internet traffic 
is colossal and will only increase in the foreseeable future~\cite{cisco2018cisco}. 
Since image compression is at the core of all video coding approaches,  improvements concerning image data are expected to have a significant impact on video as well. 
In the recent years, several neural network-based approaches 
for image 
compression~\cite{balle2017endtoend,mentzer2018conditional,balle2018full,toderici2015variable,toderici2017full} 
have been developed and rapidly caught up with several decades of work in transform coding. They are already able to outperform traditional image compression codecs, which rely on handcrafting the individual components. Instead, these methods can be trained end-to-end and leverage large amounts of data to learn an optimal non-linear transform, along with the probabilities required for entropy coding the latent representation into a compact bit stream.

While previous work has mainly focused on more efficient architectures and predictive models, in this work, we adopt a different approach by optimizing 
the latent representation \emph{individually}, 
on a per-image basis, during the encoding process. 
Thanks to this per-image adaptation, our refined representation is more 
efficient in terms of rate-distortion performance compared to the 
latent representation obtained with a simple forward pass through 
the autoencoder. 
The method is general and, as such, can be applied to improve a number of different architectures for learned image compression. 
The key benefit of the proposed solution lies in the ability to achieve an improved compression performance while the neural compression network and the predictive model are kept fixed and the computing time on the decoder side remains unchanged. 
We demonstrate the general applicability of the adaptation scheme
by providing results on two different image compression architectures~(Fig.~\ref{fig:teaser}).
The second contribution is a detailed evaluation on the use of the proposed adaptation scheme for adapting a given pre-trained model to other content that is different in visual appearance or resolution. Our evaluation includes comparisons to strategies that update only the predictive model or both the network parameters and the predictive model.  Experiments show the advantages of the latent space adaptation, as this largely allows to close the gap 
with the models specifically trained on the new content 
and does not require to transmit any updated model parameters.

\section{Preliminaries and Related Work}
The objective of \emph{lossy} image compression is to find a mapping or encoding function $\psi : \mathcal{X} \rightarrow \mathcal{Y}$
from the image space $\mathcal{X}$ to a latent space representation  
$\mathcal{Y}$ and its reverse mapping or decoding function $\phi : \mathcal{Y} \rightarrow \mathcal{X}$
back to the original image space,
with the competing constraints that, on the one hand, the latent representation 
should occupy as little storage as possible while, on the other hand, the reconstructed image should closely resemble the original image. 

In neural image compression, this mapping is realized with a neural 
encoder-decoder pair, where the bottleneck values constitute the latent 
representation. An image $x$ is first mapped to its latent 
representation $y =\psi(x)$.
After quantization, the resulting latents $\hat{y}$ are coded
losslessly to a bit stream that can be decoded into 
the image $\hat{x} = \phi(\hat{y})$.

Image compression can be formally expressed as the minimization of both the expected length of the bitstream, as well as the expected distortion of the reconstructed image compared to the original, which leads to the optimization of the following rate-distortion trade-off: 
\begin{equation}
\label{eqn:rd}
L(\psi, \phi,p_{\hat{y}}) =
\mathbb{E}_{x \sim p_x}
[\underbrace{- \log_2 p_{\hat{y}}(\hat{y})}_\text{rate}
+ \lambda~\underbrace{d(x, \hat{x})}_\text{distortion}~]\;.       
\end{equation}
Here, $d(x, \hat{x})$ is the distortion measure, 
e.g. mean squared error.  
The rate corresponds to the length of the bitstream needed to encode the 
quantized representation $\hat{y}$, based on a learned entropy model 
$p_{\hat{y}}$ over the unknown distribution of natural images $p_x$.
The weight $\lambda$ steers the rate distortion trade-off, e.g. reducing $\lambda$ leads to a higher compression rate at the cost of a larger distortion of the reconstructed image. 

In order to achieve good compression results that can deal with a vast distribution of images, two main problems arise: First, finding a powerful encoder/decoder transformation and second, properly modeling the distribution in the latent space.

Existing works such as~\cite{balle2017endtoend,toderici2015variable} have made contributions to the first problem by proposing neural network architectures to parameterize the encoding and decoding functions; more recently, the main focus of the research community has been on the second problem which not only allows to capture remaining redundancy in the latent representation for efficient entropy coding but also regularizes the encoder~\cite{balle2018variational,mentzer2018conditional,balle2018full}.

\section{Content Adaptive Compression}
In existing approaches, equation \ref{eqn:rd} is optimized over a corpus of potentially millions of images in order to find optimal functions 
for encoding and decoding ($\phi$ and $\psi$), along with a suitable probability model $p_{\hat{y}}$ for the latent space. 
Although the network has been trained over a large corpus of images to find 
what should ideally be an optimal encoding function $\psi$ over the whole data set, the encoding can still be improved by adapting to each single image.
In our work, we perform this per-image adaptation, without changing the encoder/decoder or the parameters of the latent space probability model, but by changing the latent values themselves at test time. As such, we are effectively trying to solve the following optimization problem, during test time, for a single image $x$:
\begin{equation}
\argmin_{\hat{y}} \; - \log_2 p_{\hat{y}}(\hat{y})
+ \lambda~d(x, \hat{x})\;.
\end{equation}
The fact that we do not change the decoder and the probability model in the optimization respects the assumption that both have been trained and deployed at the receiver. Therefore, the ideal strategy at this point is to find the best discrete latent representation by varying only the latent values themselves. 

There are several options to practically solve this problem, including both discrete and continuous optimization approaches. In this work, we solve it through an iterative procedure, similar as during training, where gradient descent is applied on the latents according to
\begin{equation}
y_{t+1} = y_{t} - \eta \nabla_{y} L(\psi, \phi,p_{\hat{y}}, x).
\end{equation}
Here $L(\psi, \phi, p_{\hat{y}}, x)$ is 
the rate-distortion objective for a particular image $x$:
\begin{equation}
L(\psi, \phi,p_{\hat{y}}, x) = \log_2 p_{\hat{y}}(\hat{y})
 + \lambda~d(x, \hat{x}),
\end{equation}
and $\eta$ is the weighting applied on the gradient. 
This requires a differentiable approximation of the 
quantization operation performed in the bottleneck
and we use additive uniform noise for this purpose~\cite{balle2017endtoend}.
Adopting the notation $\mathcal{U}$ for an independent uniform 
noise of width $1$, the density function $p_{\tilde{y}}$ of the random variable 
$\tilde{y} = y \, + \, \mathcal{U}(-\frac{1}{2},\frac{1}{2})$ becomes a 
continuous differentiable relaxation of the probability mass 
function $p_{\hat{y}}$.

The final image compression pipeline is described by 
Algorithm~\ref{alg:refinement}.
The lossless arithmetic encoding/decoding operations are represented by AE/AD. 
The step function corresponds to updating the latent 
representation according to the gradient
step obtained from the Adam~\cite{kingma2014adam} 
optimizer with a learning rate of $1e^{-3}$.
In all our experiments there were no noticeable improvements after
$1500$ update steps and we used this as maximum number of iterations
in Algorithm~\ref{alg:refinement}. 
On a Titan Xp GPU with 12Gb of memory, optimizing the latents 
for an HD image requires approximately 5min.

\makeatletter
\def\BState{\State\hskip-\ALG@thistlm}
\makeatother

\begin{algorithm}[h]
	\caption{Compression with Per Image Adaptation}\label{adaptlatent}
	\begin{algorithmic}[1]
		\Procedure{RefineLatents}{$y$, $x$}
		\vspace{3pt}
		\BState $\mathrm{loop\; for\; \textrm{MAX} \;steps}$:
		\State $ \tilde{y} := y + \mathcal{U}(-\tfrac{1}{2},\tfrac{1}{2})$
		\State $\tilde{x} := \phi(\tilde{y})$
		\State $L(\tilde{y}) := \sum_{i}  - \log_2 p_{\tilde{Y_i}}(\tilde{y_i}) + \lambda d(x, \tilde{x})$
		\State $y := y + \mathrm{step}(L(\psi, \phi,p_{\hat{y}}, x))$
		\BState $\mathrm{return}:$
		\State $ y$
		\EndProcedure
		
		\Procedure{Encode}{$x$}
		\State $y := \psi(x)$
		
		\State $y := \mathrm{RefineLatents}(y, x)$
		\State $\hat{y} := \mathrm{quantize}(y)$
		\State $b := \mathrm{AE}(\hat{y})$
		\BState $\mathrm{return}:$
		\State $b$
		\EndProcedure
		
		\Procedure{Decode}{$b$}
		\State $\hat{y} = AD(b)$
		\State $\hat{x} = \phi(\hat{y})$
		\EndProcedure
	\end{algorithmic}
	\label{alg:refinement}
\end{algorithm}

\section{Experimental Results}
In order to show the benefits of the proposed latent adaptation 
strategy, we consider two experimental setups: First, we explore the applicability of the proposed approach
on different image compression architectures. Second,
we evaluate how our per-image adaptation compares to other forms of content adaptation.

\subsection{Latent adaptation on different architectures}
\begin{figure}[t]
	\includegraphics[width=\columnwidth]{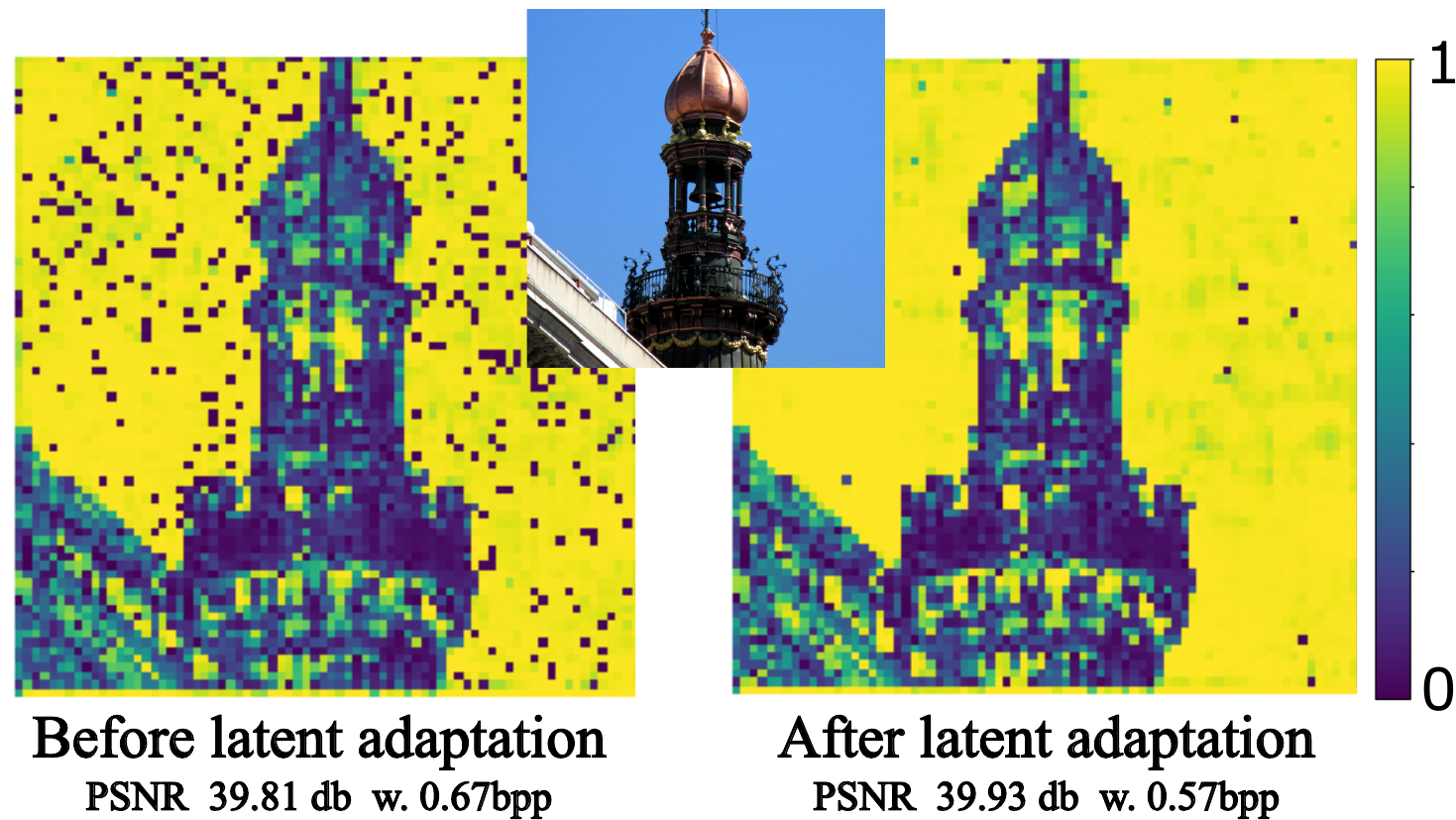}
	\caption{\textbf{Latent space adaptation.} Visualization
		of latent space likelihoods for one channel after
		adaptation. 
	\vspace{-0.4cm}
	}
	\label{fig:latents}
\end{figure}

The image-adaptive optimization is independent 
of the particular neural compression algorithm used. 
To demonstrate this, we use two existing architectures;
A (\emph{simple}) model
using a single latent space and a factorized 
probability model~\cite{balle2017endtoend},
and a more complex model (\emph{full}) with a 
hierarchical latent representation in which hyperpriors and context
are used for modeling the probabilities of the 
latents~\cite{balle2018full}. 
Both models were trained on images from the COCO segmentation 
dataset~\cite{coco}. After training, the models were tested
on the Tecnick dataset~\cite{tecnick}. 
Figure \ref{fig:teaser} showcases the rate distortion 
performance averaged over the test set together with 
JPEG2000~\cite{jpeg2000}, 
WEBP~\cite{WebPSpec} and BPG~\cite{BPGSpec} rate-distortion curves. 
As the rate-distortion curves show, in each case, the per-image 
optimized testing procedure can yield a significant increase in 
rate-distortion performance. 
Figure~\ref{fig:latents} illustrates the change in 
the likelihood of latent space values. 

\pgfplotsset{compat=1.3}
\pgfplotsset{grid style={color=black!15, },} 
\pgfplotsset{every axis x grid/.style={draw opacity=0}}

\pgfplotsset{
	axis x line=bottom,
	axis y line=left,
	axis line style={draw=none},
	tick style={draw=none},
	grid,
	xmin=0, xmax=1,
	xlabel=Rate (bpp),
	legend style={at={(0.96,0.04)},anchor=south east},
	legend cell align={left},
	label style={font=\tiny},
	legend style={font=\tiny},
	legend style={draw=gray},
	tick label style={font=\tiny},
	width=0.725\columnwidth,
	legend style = {font = {\fontsize{12 pt}{12 pt}\selectfont}},  
}
\pgfplotsset{
	every axis plot/.append style={line width=1pt, cap=round, smooth},
}
\pgfplotsset{
	compat=newest,
	/pgfplots/legend image code/.code={%
		\draw[mark repeat=2,mark phase=2,#1] 
		plot coordinates {
			(0cm,0cm) 
			(0.2cm,0cm)
			(0.4cm,0cm)
		};
	},
}

\definecolor{myBlue}{RGB}{41,77,165}
\definecolor{myBlueLight}{RGB}{164,185,226}
\definecolor{myOrange}{RGB}{231,116,46 }
\definecolor{myOrangeLight}{RGB}{239,177,105}
\definecolor{myGreen}{RGB}{78,145,41}
\definecolor{myGreenLight}{RGB}{160,219,207}
\definecolor{myRed}{RGB}{181,45,38}
\definecolor{myRedLight}{RGB}{233,138,133}
\definecolor{myMagenta}{RGB}{255,102,255}
\definecolor{myYellow}{RGB}{255,255,72}

\newcommand{\lucid}{myGreen}
\newcommand{\meridian}{myOrange}
\newcommand{\netflixFOURK}{myYellow}
\newcommand{\netflixhd}{myMagenta}


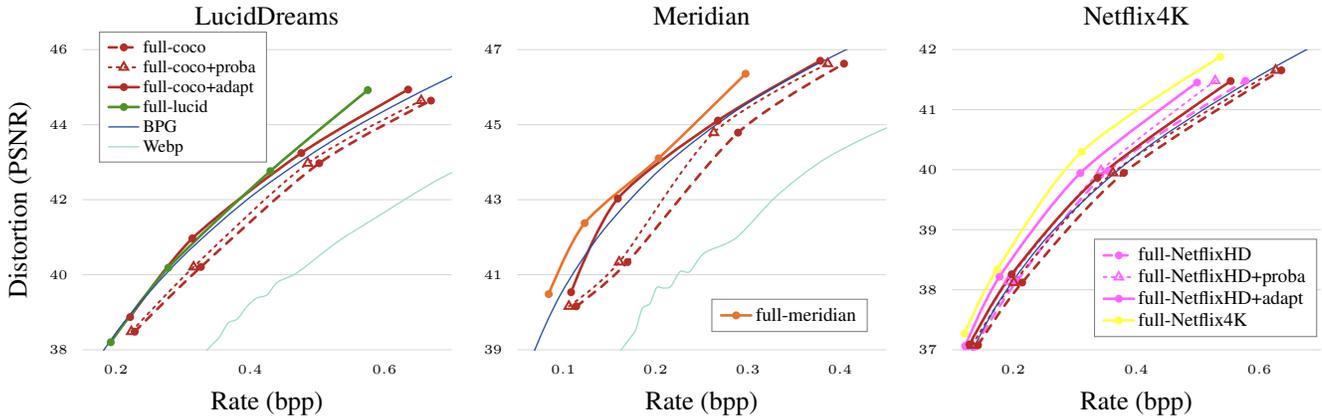
\begin{figure*}[t]
	\hspace{-0.25cm}
	\subfloat{
		\begin{tikzpicture}
		\begin{axis}[title=LucidDreams, ylabel=Distortion (PSNR),ymin=38, ymax=46,xmin=0.15, xmax=0.7,ytick={38,40,...,44,46}, scale=1.1, legend style={at={(0.5,1.06)},anchor=north east,nodes={scale=0.55, transform shape}, row sep=0.0pt} ]
		\addplot[draw=\full, dashed, mark=*, mark options={scale=0.5,\full,solid}] table [x=bpp, y=psnr, col sep=comma] {evaluation/LucidDreams_HD/full_coco.csv};
		\addlegendentry{full-coco}
		\addplot[draw=\full, dotted, line width=0.75pt, mark=triangle, mark options={scale=0.95,\full,solid}] table [x=bpp, y=psnr, col sep=comma] {evaluation/LucidDreams_HD/full_coco+proba.csv};
		\addlegendentry{full-coco+proba }
		\addplot[draw=\full, mark=*, mark options={scale=0.5,\full,solid}] table [x=bpp, y=psnr, col sep=comma] {evaluation/LucidDreams_HD/full_coco+adapt.csv};
		\addlegendentry{full-coco+adapt}
		\addplot[draw=\lucid, mark=*, mark options={scale=0.5,\lucid,solid}] table [x=bpp, y=psnr, col sep=comma] {evaluation/LucidDreams_HD/full_luciddreams.csv};
		\addlegendentry{full-lucid}
		\addplot[draw=\bpg, line width=0.5pt] table[x=bpp, y=psnr, col sep=comma] {evaluation/LucidDreams_HD/bpg444.csv};
		\addlegendentry{BPG}
		\addplot[draw=\webp, line width=0.5pt] table [x=bpp, y=psnr, col sep=comma] {evaluation/LucidDreams_HD/webp.csv};
		\addlegendentry{Webp}
		\end{axis}
		\end{tikzpicture}	
	}~
	\subfloat{
	\begin{tikzpicture}
	\begin{axis}[title=Meridian,ymin=39, ymax=47,xmin=0.05, xmax=0.45,ytick={39,41,...,47}, xtick={0.1, 0.2, ..., 0.4}, scale=1.1, legend style={at={(0.95,.17)},anchor=north east,nodes={scale=0.6, transform shape}, row sep=0.0pt} ]
	\addplot[draw=\meridian, mark=*, mark options={scale=0.5,\meridian,solid}] table [x=bpp, y=psnr, col sep=comma] {evaluation/Meridian_HD/full_meridian.csv};
	\addlegendentry{full-meridian}
	\addplot[draw=\full, dashed, mark=*, mark options={scale=0.5,\full,solid}] table [x=bpp, y=psnr, col sep=comma] {evaluation/Meridian_HD/full_coco.csv};
	\addplot[draw=\full, dotted, line width=0.75pt, mark=triangle, mark options={scale=0.95,\full,solid}] table [x=bpp, y=psnr, col sep=comma] {evaluation/Meridian_HD/full_coco+proba.csv};
	\addplot[draw=\full, mark=*, mark options={scale=0.5,\full,solid}] table [x=bpp, y=psnr, col sep=comma] {evaluation/Meridian_HD/full_coco+adapt.csv};
	\addplot[draw=\bpg, line width=0.5pt] table[x=bpp, y=psnr, col sep=comma] {evaluation/Meridian_HD/bpg444.csv};
	\addplot[draw=\webp, line width=0.5pt] table [x=bpp, y=psnr, col sep=comma] {evaluation/Meridian_HD/webp.csv};
	\end{axis}
	\end{tikzpicture}
}~
	\subfloat{
	\begin{tikzpicture}
	\begin{axis}[title=Netflix4K,ymin=37, ymax=42,xmin=0.1, xmax=0.7,ytick={36, 37,...,47}, scale=1.1, legend style={at={(0.99,.37)},anchor=north east,nodes={scale=0.6, transform shape}, row sep=0.0pt} ]
	\addplot[draw=\netflixhd, dashed, mark=*, mark options={scale=0.5,\netflixhd,solid}] table [x=bpp, y=psnr, col sep=comma] {evaluation/Netflix_4k/full_netflixHD.csv};
	\addlegendentry{full-NetflixHD}
	\addplot[draw=\netflixhd, dotted, line width=0.75pt, mark=triangle, mark options={scale=0.95,\netflixhd,solid}] table [x=bpp, y=psnr, col sep=comma] {evaluation/Netflix_4k/full_netflixHD+proba.csv};
	\addlegendentry{full-NetflixHD+proba }
	\addplot[draw=\netflixhd, mark=*, mark options={scale=0.5,\netflixhd,solid}] table [x=bpp, y=psnr, col sep=comma] {evaluation/Netflix_4k/full_netflixHD+adapt.csv};
	\addlegendentry{full-NetflixHD+adapt}
	\addplot[draw=\netflixFOURK,  mark=*, mark options={scale=0.5,\netflixFOURK,solid}] table [x=bpp, y=psnr, col sep=comma] {evaluation/Netflix_4k/full_netflix4k.csv};
	\addlegendentry{full-Netflix4K}
	\addplot[draw=\full, dashed, mark=*, mark options={scale=0.5,\full,solid}] table [x=bpp, y=psnr, col sep=comma] {evaluation/Netflix_4k/full_coco.csv};
	\addplot[draw=\full, dotted, line width=0.75pt,  mark=triangle, mark options={scale=0.95,\full,solid}] table [x=bpp, y=psnr, col sep=comma] {evaluation/Netflix_4k/full_coco+proba.csv};
	\addplot[draw=\full,  mark=*, mark options={scale=0.5,\full,solid}] table [x=bpp, y=psnr, col sep=comma] {evaluation/Netflix_4k/full_coco+adapt.csv};
	\addplot[draw=\bpg, line width=0.5pt] table[x=bpp, y=psnr, col sep=comma] {evaluation/Netflix_4k/bpg444.csv};
	\end{axis}
	\end{tikzpicture}	
	}
	\caption{\textbf{Quantitative evaluation.} 
		We compare the effect of various adaptation 
		strategies in different experimental setups (from left to right),
		by adapting to different content and resolutions (see text for details).
	}
	\label{fig:final_evaluation}
	\vspace{-0.2cm}
\end{figure*}

\subsection{Latent adaptation vs Model retraining}
In order to evaluate the benefit of an image-specific 
adaptation, we perform a comparative study with other forms
of adaptation. 
A first option consists of retraining the \emph{entire} model 
on the particular data to encode. 
Another option is to only update the probability model during the retraining procedure, i.e. to fix the feature extraction part of the network while refining the probability model on the content we wish to compress. 
From a practical point of view, these options are problematic
as additional model updates have to be transmitted whereas the latent refinement offers a way of investing more capacity into the encoding process while all model parameters remain unchanged.
In the following, we compare rate-distortion performance 
of latent adaptation with these two model retraining options
without taking into account the cost of transmitting their updated 
weights. Our objective is to get some insights regarding the capabilities 
of latent adaptation. In its current form, model retraining is not a realistic
alternative.
Next, we evaluate adaptation in two scenarios;
first in the case of content that is different in visual appearance and
then on content that is different in resolution.
For the remainder of the experiments,
we will only consider the \emph{full} model~\cite{balle2018full}.

\paragraph{Adapting to different visual appearance.}
In this experiment, we use two short movies with very different visual
appearance:
\emph{Lucid Dreams}~\footnote{https://www.youtube.com/watch?v=3zfV0Y7rwoQ}
and 
\emph{Meridian}~\footnote{https://www.netflix.com/title/80141336}.
This allows to experiment two different cases in terms of content correlation. 
In \emph{Lucid Dreams}, the frames share some similarity 
in terms of environment and characters, but the correlation is stronger 
in the second movie given the style and the lower number of scenes.
In both cases, the videos are resized to a resolution of $1280\times720$.
A common video streaming configuration is to use a key-frame
every $2$ seconds, allowing for robustness and adaptability
to the network speed. As a result, in each case, our image test set consists of $70$ frames extracted with regular spacing.
Using the full model trained on the COCO dataset, we obtain
the rate-distortion curve (dotted red curve) in
figure~\ref{fig:final_evaluation}.
This constitutes our base result. Next, we describe the methodology used for 
comparing the different adaptation strategies.

From each movie we have extracted around $1200$ frames 
(including the test frames). In a first setup, these images 
are used to train a compression network from scratch. 
The corresponding models are named 
\emph{full-lucid} and \emph{full-meridian}. 
In a second setup, the training is limited to the probability model.
Starting from the pre-trained model, we only refine the probability 
model (hyper-encoder and hyper-decoder) using the new training sets. 
This is indicated in the model name. For example, 
in the LucidDream test, the model
\emph{full-coco+proba} corresponds to a model pre-trained
on the COCO dataset for which the latent probability model 
was fine-tuned on \emph{Lucid Dreams} test images. 
Finally, we apply our iterative algorithm to adapt the latent representation 
of each image while keeping the original compression network unchanged.
We can see that, in both cases, adapting the latents
on a per-image basis always outperforms fine-tuning 
the probability model. On the \emph{Lucid Dreams} sequence, the latent adaptation even reaches the quality of the network
specialized for this sequence. 
Given the stronger correlation on the \emph{Meridian} frames, 
the specialized network performs better but adapting 
the latents still represents a significant improvement. 
\paragraph{Adapting to different resolution.} 
The objective of this experimental setup is to
obtain insights regarding the behavior of models trained for different resolutions in terms of probability 
models and features.
We extract a small set of $4$K videos from 
the (\emph{Netflix-4K}) collection, referenced by 
xiph.org~\footnote{https://media.xiph.org/video/derf/}.
On average, 20 frames per video are used as training data.
For the test set, only $2$ frames per video are used. In total, there are 25 test frames and 500 training frames.
The model trained on the COCO dataset (\emph{full-coco})
is tested first.
We then explore adaptation, first by fine-tuning 
the probability model on the $4$K training set,
and second by adapting the latents on a per-image basis.
To single out the effect of resolution, we train a new model from scratch,
\emph{full-NetflixHD}, using the frames from the set \emph{Netflix-4K}
resized to HD resolution. Then we similarly 
test fine-tuning the probability model and adapting the latents. 
Consistent with our previous experiments, adapting the latents
always outperforms fine-tuning the probability model 
and represent a significant improvement (Fig.~\ref{fig:final_evaluation}).

\section{Conclusion}
In this work we have investigated content adaptive compression strategies which can be seen as a complementary approach of improving neural image coding besides architecture refinements. 
More specifically, we have presented a latent space refinement algorithm that allows to improve quality by roughly 0.5 dB at the same bit rate on the Tecnick data set.
This strategy also allows to significantly close the gap between generic pre-trained models and models that are specifically trained for a given target content.
Thus, the latent space adaptation can be an effective strategy to make a given encoding process more powerful and content adaptive. This is particularly beneficial in situations such as streaming, where the encoding complexity is not the limiting factor when compared to the transmission and decoding. 
As the gap towards models that are \emph{entirely} trained on the specific target content cannot fully be closed, it would be interesting to further investigate which more complex but still practically viable form of adaptation may achieve this. Also currently, neural image compression models are typically trained for each rate-distortion point and it would be similarly beneficial to investigate strategies that allow automatic adaptation to each quality level.

{\small
	\bibliographystyle{ieee}
	\bibliography{compression}
}

\end{document}